\documentclass[conference]{IEEEtran}
\def\IEEEtitletopspace{18pt}
\usepackage[numbers]{natbib}

\usepackage{amsmath,amssymb,amsfonts}
\usepackage{algorithmic}
\usepackage{graphicx}
\usepackage[hidelinks]{hyperref}
\usepackage{textcomp}
\usepackage{booktabs}
\usepackage{subcaption}
\usepackage{xcolor}
\def\BibTeX{{\rm B\kern-.05em{\sc i\kern-.025em b}\kern-.08em
    T\kern-.1667em\lower.7ex\hbox{E}\kern-.125emX}}
\usepackage{flushend}
    
\usepackage{pgfplots}

\pgfplotsset{compat=1.17}
\usetikzlibrary{pgfplots.groupplots}

\newcommand{\cvec}[1]{\boldsymbol{\mathrm{#1}}}

\newcommand{\p}{p}
\newcommand{\mpw}{{\cvec{\omega}}}
\newcommand{\dtw}{{\nu}}
\newcommand{\dtwvec}{{\cvec{\dtw}}}
\newcommand{\mpp}{{\cvec{\theta}}}
\newcommand{\Mpp}{{\cvec{\Theta}}}
\newcommand{\ctxp}{{\cvec{\eta}}}
\newcommand{\optp}{{\cvec{\lambda}}}

\newcommand{\policy}[0]{\p_\Mpp(\mpw|\cvec c)}

\newcommand{\gating}[1]{{#1}(o|\cvec c)}

\newcommand{\mgating}[0]{\p_\optp(o)}

\newcommand{\ctxtdistr}[0]{\p(\cvec c)}
\newcommand{\ctxtcomp}[0]{\p_\ctxp(\cvec c|o)}
\newcommand{\ctxtcompdata}[0]{\p_\ctxp(\cvec c_i|o)}

\newcommand{\comppi}[0]{\p_\mpp(\mpw | \cvec c, o)}
\newcommand{\comppidata}[0]{\p_\mpp(\mpw_i|\cvec c_i, o)}

\newcommand{\dataweight}[0]{\dtwvec}
\newcommand{\idataweight}[0]{\dtw_i}

\newcommand{\idatacompweight}[0]{ \dtw_{o, i}}
\newcommand{\idataresponsibility}[0]{\tilde{\dtw}_{o|i}}

\newcommand{\objref}[1]{\hyperref[{#1}]{Objective~\ref*{#1}}}

\begin{document}

\title{Curriculum-Based Imitation of Versatile Skills
}

\author{
\IEEEauthorblockN{Maximilian Xiling Li\IEEEauthorrefmark{1}\IEEEauthorrefmark{3},
Onur Celik\IEEEauthorrefmark{2}, Philipp Becker\IEEEauthorrefmark{2}, Denis Blessing\IEEEauthorrefmark{2}, Rudolf 
Lioutikov\IEEEauthorrefmark{1}, Gerhard Neumann\IEEEauthorrefmark{2}} 
\IEEEauthorblockA{\IEEEauthorrefmark{1} Intuitive Robots Lab, Karlsruhe Institute of Technology, Karlsruhe, Germany} 
\IEEEauthorblockA{\IEEEauthorrefmark{2} Autonomous Learning Robots, Karlsruhe Institute of Technology, Karlsruhe, Germany}
}
\maketitle
\begingroup\renewcommand\thefootnote{\IEEEauthorrefmark{3}}
\footnotetext{Correspondence to \texttt{maximilian.li@kit.edu}}
\endgroup

\begin{abstract}
Learning skills by imitation is a promising concept for the intuitive teaching of robots.  
A common way to learn such skills is to learn a parametric model by maximizing the likelihood given the demonstrations.
Yet, human demonstrations are often multi-modal, i.e., the same task is solved in multiple ways which is a major challenge for most imitation learning methods that are based on such a maximum likelihood (ML) objective.
The ML objective forces the model to cover all data, it prevents specialization in the context space and can cause mode-averaging in the behavior space, leading to suboptimal or potentially catastrophic behavior. 
Here, we alleviate those issues by introducing a curriculum using a weight for each data point, allowing the model to specialize on data it can represent while incentivizing it to cover as much data as possible by an entropy bonus.
We extend our algorithm to a Mixture of (linear) Experts (MoE) such that the single components can specialize on local context regions, while the MoE covers all data points. 
We evaluate our approach in complex simulated and real robot control tasks and show it learns from versatile human demonstrations and significantly outperforms current SOTA methods. \footnote{A reference implementation can be found at \\\href{https://github.com/intuitive-robots/ml-cur}{\texttt{https://github.com/intuitive-robots/ml-cur}}}
\end{abstract}


\section{Introduction}
To increase the accessibility of robotic systems, we need intuitive ways to teach them new skills.
One promising and well-studied approach to this problem is imitation learning. 
In this setting, the robot learns new skills from demonstrations provided by a domain expert, e.g., through teleoperation. 
Most imitation learning approaches then fit a simple probabilistic model to those demonstrations allowing them to copy the expert's behavior by sampling from the model.

Learning a skill model from real word demonstrations commonly suffers from three significant problems, \textbf{a)} \textit{outlier-sensitivity}, \textbf{b)} \textit{locality-violations} in context space and \textbf{c)} \textit{mode-averaging} in behavior space.
Consider kinesthetic demonstrations of a robot hitting a table tennis ball.
Commonly, some demonstrations will run into joint limits, however, these demonstrations might be few and significantly different from other demonstrations.
In common approaches, such demonstrations have to be filtered out manually a priori or the learned model will include such degenerating demonstrations into its behavior, suffering from \textit{outlier-sensitivity} \citep{mandlekar2021what}.
Furthermore, the context points in such an example would cover a large area of the table.
Covering disjoint or excessively large areas of the context space generally leads to bad behavior, especially if the skill parametrization is linear in the context, which is a common assumption due to its computational simplicity.
Avoiding this \textit{locality-violation} allows each skill to specialize on a local subset of the contexts-space, performing better in that particular region than an overly general skill.
Finally, significantly different strokes that return the ball to the same point are an example of the multi-modality generally found in real-world demonstrations \citep{orsini2021}.
Common approaches combine these different modes into one interpolated, suboptimal behavior that suffers from \textit{mode-averaging}.
If the number of components is less than the unknown number of modes present in the demonstrations, this problem even exists for mixture models that are optimized using common approaches such as EM and other likelihood-based methods. 

We propose a novel approach to learning a Mixture of Experts (MoE) models that avoid these three problems. 
To this end, we introduce \textbf{a)} per-component weights on each data point that allow each component to select its own curriculum, i.e., the subset of data points it considers for its behavior. 
This weighting explicitly allows each component to ignore unfavored data points, avoiding \textit{outlier-sensitivity}.
\textbf{b)} a per-component Gaussian context distribution that prevents \textit{locality-violations} in the context space and allows each component to learn specialized behavior for a region of the context space. And \textbf{c)} an optimization objective for updating the weights that prevent each component from \textit{mode-averaging}, by learning the weights that optimize the likelihood and the context distribution simultaneously, while still incentivizing a larger coverage of data points through an entropy bonus.

\begin{figure*}[t]
    \centering
    \includegraphics[width=0.9\textwidth]{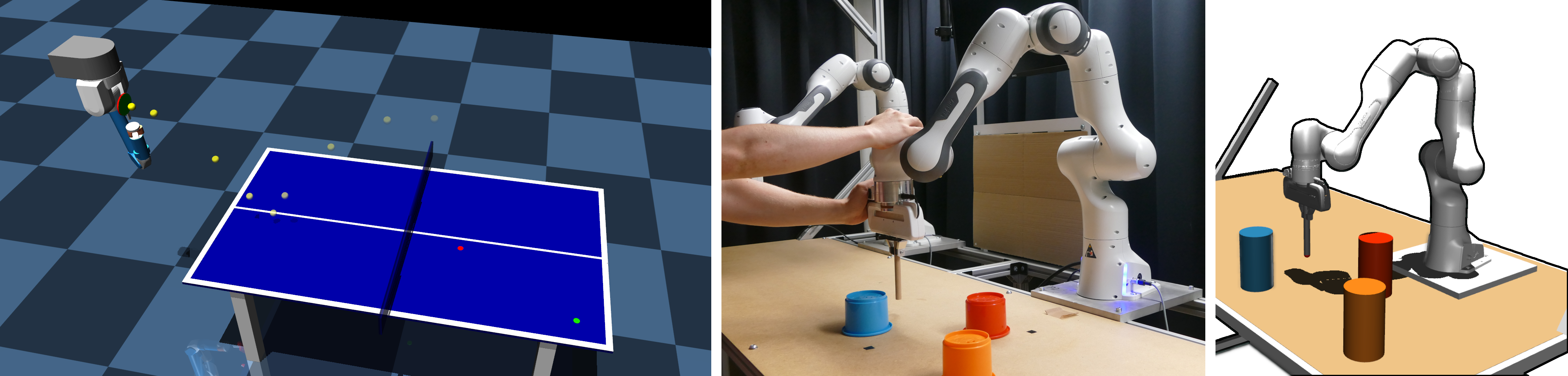}
    \caption{\textbf{Left:} Virtual Table Tennis environment. The ball is launched above the red dot towards the robot and is supposed to hit on the green dot on return. 
    \textbf{Center:} Data Collection for the obstacle avoidance task by a human demonstrator.   
    \textbf{Right}: Virtual twin for the obstacle avoidance task. 
    For each demonstration the obstacles, serving as contexts, are randomly placed.}
    \label{fig:setup}
\end{figure*}

We show our approaches' advantages in three experiments and a detailed ablation study. 
For the experiments, we use a simple toy example, a sophisticated table-tennis simulation, and a highly multi-modal, contextual obstacle avoidance task using a Franka Emika Panda 7DoF Manipulator, as shown in \autoref{fig:setup}.
For the real robot task, we use versatile human demonstrations that exhibit the above-mentioned challenging properties.
We compare our method to several baselines, i.e., Expectation Maximization \citep{dempster1977maximum}, Expected Information Maximization \citep{becker2020expected}, K-Nearest Neighbours, Normalizing Autoregressive Flows \citep{huang2018neural}, and a Mixture Density Network-based approach \citep{zhou2020movement}.  We outperform the baselines in all tasks.

\section{Preliminaries}\label{sec::preliminaries}
We consider a given set of demonstrations $\mathcal{T}=\lbrace\left(\cvec{\tau}_i, \cvec{c}_i\right) \rbrace_{i=1}^{N}$. A trajectory demonstration $\cvec{\tau}_i = \left(\cvec{s}_0, \cvec{s}_1, ... \cvec{s}_{T_i}\right)$ consists of the recorded states, e.g., joint configurations, $\cvec{s}_t$ of the robot and the context $\cvec{c}_i$ defines task properties required to contextualize a given demonstration.  
The context vector $\cvec{c}$ encodes environmental states and desired goal conditions, such as obstacle or target positions. In contrast to prior work \citep{seker2019conditional, Calinon2015tpgmm, le2021forceful} the context vector is not required to be in the sensorimotor space of the robot nor does it specify a new coordinate system. 
The demonstration length $T_i$ is different for each $\cvec{\tau}_i$.
We represent skills as movement primitives~\cite{schaal2005learning}, specifically Probabilistic Movement Primitives (ProMPs)~\cite{paraschos2013probabilistic}.
ProMPs use time-dependent radial basis functions $\cvec{\Phi}$, that are centered at different points in time and linearly combined with a parameter vector $\mpw$ to represent trajectories. 
This compactly represents each expert trajectory as a parameter vector $\mpw$ which is fitted to the demonstration using linear regression
\begin{align}\label{eq:ProMPReg}
    \mpw_i = \left(\cvec{\Phi}^T\cvec{\Phi}+\lambda \cvec{I}\right)^{-1}\cvec{\Phi}^T\cvec{\tau}_i.
\end{align}
Exact details are shown in \cite{paraschos2013probabilistic}.
Together with the context for each demonstration, we obtain a data set $\mathcal{D}=\lbrace \cvec{c}_i, \mpw_i \rbrace_{i=1}^{N}$ which we can use to fit our model. 
To capture the versatility, we employ a Mixture of Expert (MoE) model, given as 
\begin{align}\label{eq::standard_mixture_model}
    p(\mpw | \cvec{c}) = \sum_o p(\mpw | \cvec{c}, o) p(o | \cvec{c}).
\end{align}
Here $p(o | \cvec{c})$ is the softmax gating distribution and $p(\mpw | \cvec{c}, o)$ are linear conditional Gaussian experts.

\section{Related Work}

\textbf{Imitation Learning for Versatile Skills.}
In the simplest case, learning skills as movement primitives \cite{schaal2005learning, paraschos2013probabilistic} from human demonstrations amounts to regression.
To capture versatility in the behavior we need to learn multi-modal models, e.g., mixture models.
Common approaches that learn GMMs over MP parameters via EM, only use the resulting mixture models to find suitable MP parameters for novel contexts \citep{mulling2013learning, ewerton2015learning}.
Neither of these approaches capture versatile behavior for individual contexts. Other approaches learn a mixture over contextualized MPs using Gaussian Mixture Regression  \citep{pervez2018learning} and learn 
non-linear relations between context and MP parameters by using a mixture density network \citep{zhou2020movement}. 
All of the above-mentioned approaches use maximum likelihood objectives and thus suffer from the mode-averaging problem.
Expected Information Maximization (EIM)\cite{becker2020expected} computes the information projection \cite{murphy2012machine} based on samples that can be applied to skill-based imitation learning \cite{freymuth2021versatile}. 
Yet, EIM relies on an intermediate density ratio estimation step \cite{sugiyama2012density}, making the approach inefficient and hard to tune. EIM further suffers from \textit{locality-violations}.
Recent approaches use generative deep learning models \cite{seker2019conditional, hausman2017multi} or apply preference learning using additional human rankings \citep{myers2022learning}. However, these approaches do not leverage skill-based representations but instead operate on entire trajectories of state-action pairs.

\textbf{Curriculum Learning.}
While curricula are a well-studied technique to improve the performance of supervised \cite{bengio2009curriculum, soviany2021curriculum} and reinforcement learning \cite{narvekar2020curriculum} approaches, we are not aware of any approaches using curricula for skill-based imitation learning. 
Our approach is inspired by a recent approach to versatile skill discovery in reinforcement learning \cite{celik2022specializing}. 
They employ the same MoE parametrization, as a basis for a curriculum approach to versatile maximum entropy reinforcement learning. 
Yet, as they consider a fundamentally different setting they use a different formulation of the curriculum. 
Further, they only addressed \textit{locality-violations} as reinforcement learning objectives naturally do not suffer from \textit{mode-averaging}.

\section{Curriculum-Based Imitation Learning}
Consider a given set of demonstrations $\mathcal{T}=\lbrace\left(\cvec{\tau}_i, \cvec{c}_i\right) \rbrace_{i=1}^{N}$, where each tuple consists of a demonstrated trajectory $\cvec\tau_i$ and a corresponding context variable $\cvec{c}_i$, e.g., a joint trajectory of a robot hitting a table tennis ball and the balls subsequent point of impact on the table. Our goal is now to learn a parameterized, generative contextual skill model $p_\Mpp(\mpw|\cvec{c})$ over a motion primitive weighting $\mpw$ given a particular context $\cvec c$ under the parameters $\Mpp$, e.g., predict the weights of a primitive that hits a table tennis ball such that it bounces of the table at a particular position. In this work we use the ProMP representation and, hence, apply the weight projection in \autoref{eq:ProMPReg} to transform the dataset $\mathcal{T}=\lbrace\left(\cvec{\tau}_i, \cvec{c}_i\right) \rbrace_{i=1}^{N}$ into the corresponding dataset $\mathcal{D}=\lbrace\mpw_i, \cvec{c}_i \rbrace_{i=1}^{N}$. Note that such a projection is not unique to ProMPs and other primitive representations could be used just as well.
We first motivate our objective for a single component in and then extend it to the Mixture of Experts (MoE) case. We call the resulting method \emph{Maximum-Likelihood based Curriculum (ML-Cur)}. 

\textbf{Curriculum Imitation Learning for a Single Expert.}
An intuitive objective for fitting a model is to maximize the log-likelihood  
$ \mathcal{L}\left(\mpp|\mathcal{D}\right) = \sum_i \log p_\mpp(\mpw_i|\cvec c_i)$, with $\mpp$ being the parametrization of a single component, e.g., the mean and covariance of a ProMP weight distribution. 
While easy to optimize, this approach suffers from the discussed shortcomings \textit{outlier-sensitivity}, \textit{locality-violation} and \textit{mode-averaging}. 
By introducing a weight $\dtw_i$ for each demonstration the component is able to only consider a relevant subset of the given demonstrations, i.e. its curriculum, allowing the component to ignore outliers or data points from other modes. The weights $\dtw$ sum up to 1 and are strictly positive.
We further introduce a Gaussian context distribution $p_\ctxp(\cvec c)$ restricting the component to a local subspace of the context and enables the component to learn more specialized behavior, addressing the \textit{locality-violation} problem.
The resulting objective
\begin{align}
\label{eq::GreedySingleObjective}
    \arg\max_{\dtw_{1:N}, \ctxp, \mpp} \sum_i \dtw_i \left(\log p_\mpp(\mpw_i|\cvec c_i)+ \log p_\ctxp(\cvec c_i)\right),
\end{align}
ensures that the maximum likelihood updates for $p_\ctxp(\cvec c)$ and $p_\mpp(\mpw|\cvec c)$ are \textbf{i)} coupled, \textbf{ii)} only consider data points that $p(\mpw|\cvec c)$ can represent and \textbf{iii)} the considered data points are local in the context space.
However, the objective in \autoref{eq::GreedySingleObjective} would encourage greedy components with increasingly smaller curricula. We counter this behavior by introducing an entropy bonus over the data weights $\mathcal{H}\left[\dtwvec\right] = -\sum_i \dtw_i \log \dtw_i$, yielding the single component objective
\begin{align}
\label{eq::SingleObjective}
    \arg\max_{\dtw_{1:N}, \ctxp, \mpp} & \sum_i \dtw_i \left(\log p_\mpp(\mpw_i|\cvec c_i)+ \log p_\ctxp(\cvec c_i)\right) \nonumber \\
    & + \alpha \mathcal{H}\left(\dtwvec\right),
\end{align}
with scaling factor $\alpha$. This objective yields an exciting new type of learning process which forms the core of our proposed method. At the beginning the component will focus on a small curriculum, i.e., a small set of data points, since the log-likelihood will dominate the objective. With increasing learning process the likelihood of more data points will increase and the entropy term will force the component to extend its curriculum to more data points until the `representational capacity' of the component is reached and further data points can not be represented well any more. At the same time the context distribution ensures the locality of the curriculum in the context space.

\objref{eq::SingleObjective} can now be iteratively optimized for $p_\mpp(\mpw|\cvec c)$ and $p_\ctxp\left(\cvec c\right)$ and subsequently for the data weights $\dtw_{1:N}$. The updates for $\mpp$ and $\ctxp$ follow common derivations for MoE models. 
However, the closed form solution for the data weights
\begin{align}\label{eq::weight_update_single}
    \dtw_i^* \propto \left(p_\mpp\left(\cvec \mpw_i | \cvec c_i\right) p_\ctxp\left(\cvec c_i\right) \right)^{\frac{1}{\alpha}}
\end{align}
offers the valuable insight that the weight $\dtw_i$ is high iff $p_\mpp\left(\cvec \mpw_i | \cvec c_i\right) p_\ctxp\left(\cvec c_i\right)$ is high, allowing the component to adjust its curriculum by specializing on favored $(\cvec \mpw_i, \cvec c_i)$ tuples.
Note that this specialization is only possible as we iteratively optimize the individual terms. 
This weight update in combination with the locality provided by the Gaussian context distribution further ensures that the weighted maximum likelihood updates for $\mpp$ and $\ctxp$ are not subject to mode averaging.

The scalar factor $\alpha$ in \autoref{eq::SingleObjective} trades off the importance of the likelihood terms and the entropy of the weight distribution.  
For high $\alpha$ values the updates for $p(\cvec c)$ and $p(\mpw|\cvec c)$ will approach the standard maximum likelihood update as the weights will get more and more uniform.
In contrast, a too-small $\alpha$ can cause the expert to concentrate on a single data point only. Due to highly varying ranges of the log likelihoods and the entropy values during training, choosing a good value is not straightforward.
We therefore additionally propose an extension that automatically tunes $\alpha$.

\textbf{Curriculum-Based Imitation Learning for MoE.}\label{sec::multi}
We propose a multi-modal model that allows individual components to \textbf{a)} specialize on local context regions while avoiding spanning over the whole context space, \textbf{b)} ignore modes it cannot represent, and \textbf{c)} still cover as much as possible of the available data.
Following \cite{celik2022specializing}, we define our mixture model as  
\begin{align}\label{eq::curr_mixture}
     \policy &=\sum_o \dfrac{\ctxtcomp \mgating}{\ctxtdistr} \comppi,
\end{align}
where $\comppi$ and $\ctxtcomp$ denote the component and the context distribution analogous to above, which are now additionally conditioned on the respective component.
The $K$-variate categorical distribution $\mgating$ models a gating prior over the $K$ components and $\ctxtdistr=\sum_o\ctxtcomp \mgating$ denotes a prior over the context.
The model in \autoref{eq::curr_mixture} differs from standard mixture of experts (\autoref{eq::standard_mixture_model}) by modelling the component-wise context distribution $\ctxtcomp$ and $\mgating$ instead of the gating distribution $\gating{\p}$.
This formulation defines a mixture model where each component has its own curriculum while focusing on a local context region.

A naive approach to update the MoE model is to use the objective in \autoref{eq::SingleObjective} and let each component specialize independently on a different part of the data set. In this case however, components might adjust their curriculum similarly and thus, concentrate on the same data regions, as they are not aware on which regions the other components specialize.
To allow each components to adjust its curriculum independently while coordinating the curriculum with other components, we define individual data weights $\dtw_{o,i}$ for each component, yielding the per-datapoint weight $\idataweight = \frac{1}{K}\sum_o\idatacompweight$.
In combination with the entropy bonus in \autoref{eq::SingleObjective} such weights force the components to focus on different parts of the context space.
Following the described intuitions, we extend our objective (\autoref{eq::SingleObjective}) to
\begin{align}\label{eq::SampleBasedObjective}
    \arg\max_{\Mpp} & \sum_i\sum_{o,\mpp,\ctxp} \idatacompweight(\log \comppidata + \log \ctxtcompdata \nonumber\\
    & + \log \mgating) + \alpha\mathcal{H}(\dataweight),
\end{align}
with $\mathcal{H}(\dataweight) = -\sum_i\frac{1}{K}\sum_o\idatacompweight \log \frac{1}{K}\sum_o \idatacompweight$
being the entropy over all per-component data weights and $\Mpp = \lbrace \dtw_{1:K,1:N}, \ctxp_{1:K}, \mpp_{1:K}, \optp\rbrace$ being the learned parameters of the model.
This objective is very similar to the one in Eq. (\ref{eq::SingleObjective}), but with the key distinction that the likelihood and weight terms are now conditioned on the mixture variable $o$.

\textbf{Optimizing the MoE.}\label{sec::OptiMulti}
As described and motivated above, we now seek to maximize \objref{eq::SampleBasedObjective} w.r.t. $\dtw_{1:K,1:N}, \ctxp_{1:K}$ and $\mpp_{1:K}$.
This optimization can be easily done with weighted maximum likelihood updates. However, the optimization w.r.t. $\idatacompweight$ is difficult since the sum over $o$ appears inside the log of the entropy term, rendering a closed form solution for $\dtw_{o,i}$ infeasible. 
We tackle this problem by introducing a (tight) variational lower bound similar to \cite{arenz2020trust}
\begin{align}\label{eq::LowerBound}
    \arg\max_{\Mpp} &~~ \sum_i\sum_{o,\mpp,\ctxp} \idatacompweight(\log \comppidata + \log \ctxtcompdata \nonumber\\
    & ~~ + \log \mgating) + \alpha \mathcal{H}(\idatacompweight), 
\end{align}
with $\mathcal{H}(\idatacompweight) = - \frac{1}{K}\sum_i\sum_o\idatacompweight (\log \idatacompweight - \log \idataresponsibility)$ and $\idataresponsibility = \idatacompweight^{old}/\sum_o\idatacompweight^{old}$ being the per component data entropy and component data responsibility respectively. Intuitively, the log of the data responsibility returns a high value, if only one component $o$ is responsible for data point $i$, encouraging the components to divide the data points among each other and thus avoid overlapping support. 
Note that the old weight $\idatacompweight^{old}$ is simply fixed to the weight of the previous iteration. 
The lower bound in \autoref{eq::LowerBound} does not have any effect on the weighted maximum likelihood update rules for $\ctxp_{1:K}, \mpp_{1:K}$ and $\optp$, but yields a closed-form solution for the component weight update
\begin{align}\label{eq::WeightUpdate}
    \idatacompweight \propto \left(\comppidata\ctxtcompdata\mgating\right)^{K/\alpha}\idataresponsibility,
\end{align}
recovering the solution from \autoref{eq::weight_update_single} adjusted to mixture models. 

\textbf{Autotuned Entropy Scaling.}\label{sec::autotuned_scale}
Choosing the entropy scaling parameter $\alpha$ is not straightforward and might yield unsatisfying results, caused by highly varying ranges of the log likelihoods and the entropy values during training. To stabilize the optimization for $\idatacompweight$, we propose to reformulate our objective in \autoref{eq::LowerBound} as a constrained optimization problem 
\begin{align}\label{eq::autotuning_alpha}
    \arg\max_{\Mpp} & \quad \sum_i\sum_{o,\mpp,\ctxp} \idatacompweight 
    (\log \comppidata + \log \ctxtcompdata \nonumber \\ 
    & \quad + \log \mgating + \alpha_o \log \idataresponsibility) \\ 
    \text{s.t.} &\quad
     \mathcal{H}(\idatacompweight) \geq \mathcal{H_{\text{min}}} \nonumber ,
\end{align}
where we now consider a scaling $\alpha_o$ and a minimal entropy $\mathcal{H_{\text{min}}}$ per component $o$. Choosing $\mathcal{H_{\text{min}}}$ is easier and more intuitive compared to choosing the entropy scaling factor $\alpha$ as $\mathcal{H_{\text{min}}}$ ensures that each component covers a certain amount of samples, consequently preventing components from collapsing to a single sample. A good choice is to set $\mathcal{H_{\text{min}}} =\log n_{\text{eff}}$, where $n_{\text{eff}}$ is the desired number of effective samples per component.
Furthermore, we are not restricted to choose one single $\alpha$ for all components anymore.
The Lagrangian dual optimization yields the optimal solution for each weight 
\begin{align*}
    \idatacompweight^* \propto    \left(\comppidata\ctxtcompdata\mgating\right)^{\frac{1}{\alpha_o}}\idataresponsibility,
\end{align*}
with $\alpha_o$ being obtained by minimizing the dual function 
\begin{align*}   
\footnotesize{\arg \min_{\alpha_o \geq 0}
\alpha_o  \left(
\log \sum_i \left(\comppidata\ctxtcompdata\mgating\right)^{\frac{1}{\alpha_o}}\idataresponsibility
 - \mathcal{H}_{\text{min}}  \right)}.
\end{align*}
We optimize the convex dual using L-BFGS\citep{liu1989limited}. 

\section{Experiments}
We show the advantages and strengths of our method on three experimental settings, including real-world robots and data.
The baselines include Mixture of Expert (MoE) (\autoref{eq::standard_mixture_model}) trained with both Expectation Maximization (EM)~\cite{dempster1977maximum} and Expectation Information Maximization (EIM)~\cite{becker2020expected}. 
While EM suffers from \textit{outlier-sensitivity}, \textit{locality-violations} and \textit{mode-averaging}, EIM adresses \textit{mode-averaging}
yet still suffers from \textit{locality-violations}. 
EM and ML-Cur only have a small number of hyper-parameters. EIM relies on an intermediate density ratio estimation~\cite{sugiyama2012density}, making the approach inefficient and hard to tune. We additionally compare to Mixture Density Networks (MDN)-based approach~\cite{zhou2020movement},
Neural Autoregressive Flows (NFlow)~\cite{huang2018neural} and K-nearest neighbor (KNN)~\cite{bishop2006pattern}.

\textbf{Planar Robot Reacher.} First, we consider a synthetic 2D robotic reaching task for illustrative purposes.
\begin{figure}[t]
    \centering
    \resizebox{0.48\textwidth}{!}{\input{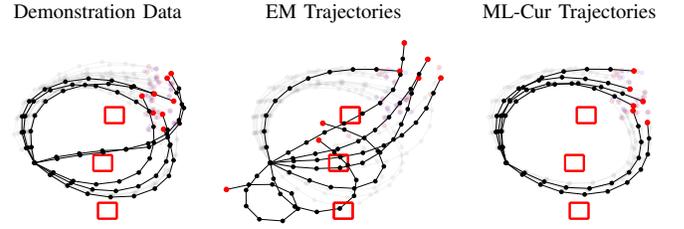}}
    \caption{\textbf{Planar Robot Reacher.} Trajectories in our planar reacher toy task. The target positions are marked in purple. 
    Black lines represent the links of a 10DoF robotic arm with the end-effector highlighted in red.
    The environment obstacles are visualized as red boxes. The demonstrations used for training (left) show three ways of reaching the target.
    We fit MoEs with 2 components with both EM (center) and ML-Cur (right). ML-Cur avoids the mode-averaging problem that EM suffers from.}
    \label{fig:planar}
\end{figure}
A 10DoF planar robot must reach a specified target position while avoiding three obstacles.
The task is contextualized by the 2D target position. The action space is a ten-dimensional joint angle position vector.
Using 1780 collision-free training demonstrations, we fit MoE models with two components using EM and ML-Cur. 
As shown in \autoref{fig:planar} EM is unable to fit the training demonstrations as the maximum likelihood objective forces the model to average over the middle and lower mode in the data, which does not only prevents the robot from reaching the targets but also makes it crash into the obstacles. ML-Cur on the other hand simply ignores the mode with the least amount of samples and yields a successful and safe model.

\textbf{Table Tennis.}
\begin{figure}[t]
\centering
\resizebox{0.48\textwidth}{!}{\begin{tikzpicture} 
\definecolor{darkgray158}{RGB}{158,158,158}
\definecolor{darkorange2551490}{RGB}{255,149,0}
\definecolor{darkslategray71}{RGB}{71,71,71}
\definecolor{lightgray204}{RGB}{204,204,204}
\definecolor{limegreen018569}{RGB}{0,185,69}
\definecolor{orangered255440}{RGB}{255,44,0}
\definecolor{skyblue102204238}{RGB}{102,204,238}
\definecolor{slategray13291151}{RGB}{132,91,151}
\definecolor{teal1293165}{RGB}{12,93,165}
    \begin{axis}[%
    hide axis,
    xmin=10,
    xmax=50,
    ymin=0,
    ymax=0.4,
    legend style={
        draw=white!15!black,
        legend cell align=left,
        legend columns=-1, 
        legend style={
            draw=none,
            column sep=1ex,
            line width=1pt
        }
    },
    ]
    \addlegendimage{teal1293165, very thick}
    \addlegendentry{EIM};
    \addlegendimage{limegreen018569, very thick}
    \addlegendentry{EM};
    \addlegendimage{darkorange2551490, very thick}
    \addlegendentry{ML-Cur};
    \addlegendimage{orangered255440, very thick}
    \addlegendentry{MDN};
    \addlegendimage{slategray13291151, very thick}
    \addlegendentry{NFlow};
    \addlegendimage{darkslategray71, very thick}
    \addlegendentry{1-NN};
    \addlegendimage{darkgray158, very thick}
    \addlegendentry{10-NN};
    \end{axis}
\end{tikzpicture}}\linebreak
\resizebox{0.48\textwidth}{!}{\input{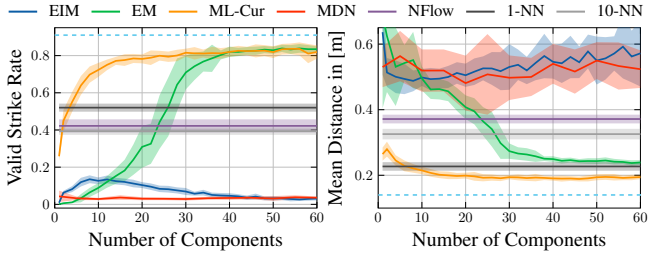}}
    \caption{\textbf{Table Tennis.}
    NFlow and KNN with $k=1$ and $k=10$ appear as straight lines, as these models don't have components. 
    All models have been tested with 20 different random seeds and training-test selections of the available data.
    ML-Cur outperforms the baselines in both the valid strike rate and MDE. It most closely approaches the performance of the demonstrator, shown as a light blue dashed line.
    }
    \label{fig:tt_algocomp}
\end{figure}
Next, we consider a virtual table tennis task modeled in MuJoCo~\cite{todorov2012mujoco}. 
We collect a data set using a highly versatile reinforcement learning agent~\cite{celik2022specializing} and randomly split it into 5,000 training and 500 test samples.
A ball is launched from different start positions, and after bouncing on the robot's table side, the robot has to return the ball to a given goal position. 
The expert demonstrations entail varying movements, such as forehand and backhand strokes. 

The context is given as a 4D vector, defining the XY ball launch position and the XY target position on the opponent's side.
We evaluated two metrics, the \emph{valid strike rate}, i.e., how often the ball is successfully returned to the other side, and the \emph{mean distance error (MDE)} to the target position for valid strikes.  
\autoref{fig:tt_algocomp} shows the metrics for ML-Cur, EM, EIM, MDN, NFlow, KNN with one (1-NN) and ten (10-NN) nearest neighbors for a varying number of model components.
ML-Cur generally outperforms EM, especially with lower numbers of components, while EIM and MDN do not seem to be able to solve the task as indicated by the low valid strike rate. 
Nflow, 1-NN, and 10-NN have a higher valid strike rate for a small number of components, but EM and especially ML-Cur outperform them with more components.

\begin{figure}[t]
    \centering
    \resizebox{0.48\textwidth}{!}{
\begin{tikzpicture}

\definecolor{darkorange2551490}{RGB}{255,149,0}
\definecolor{lightgray204}{RGB}{204,204,204}
\definecolor{skyblue102204238}{RGB}{102,204,238}

\begin{groupplot}[group style={group size=2 by 1, horizontal sep=1.5cm}]
\nextgroupplot[
tick pos=both,
width=8cm, 
height=5cm, 
xlabel={Minimum Number Effective Samples},
xmajorgrids,
xmin=-18.95, xmax=419.95,
xtick style={color=black},
ylabel={Valid Strike Rate},
ymajorgrids,
ymin=-0.0421230533399924, ymax=0.876029018943558,
ytick style={color=black}
]
\path [fill=darkorange2551490, fill opacity=0.3]
(axis cs:1,-0.000388868236194652)
--(axis cs:1,0.00558886823619465)
--(axis cs:2,0.0174300263264058)
--(axis cs:3,0.0155330532163432)
--(axis cs:4,0.0307122484231095)
--(axis cs:5,0.216947908113579)
--(axis cs:10,0.439572820181605)
--(axis cs:15,0.504078032124144)
--(axis cs:20,0.616642189532471)
--(axis cs:25,0.715242950977745)
--(axis cs:30,0.787916327062906)
--(axis cs:35,0.783942837218564)
--(axis cs:40,0.784052846164867)
--(axis cs:45,0.790480387271353)
--(axis cs:50,0.790971669233827)
--(axis cs:55,0.794533502536981)
--(axis cs:60,0.815553453345957)
--(axis cs:65,0.814019992006394)
--(axis cs:70,0.81241110789795)
--(axis cs:75,0.816467790240896)
--(axis cs:80,0.825691912636123)
--(axis cs:85,0.829531390071027)
--(axis cs:90,0.825713206001022)
--(axis cs:95,0.83429483383976)
--(axis cs:100,0.824364959911047)
--(axis cs:150,0.815839358732353)
--(axis cs:200,0.761630596076921)
--(axis cs:250,0.667656611483145)
--(axis cs:300,0.492640428100132)
--(axis cs:310,0.438411446157224)
--(axis cs:320,0.392544653440418)
--(axis cs:330,0.02574330356053)
--(axis cs:340,0.00467058484879019)
--(axis cs:350,0.00467058484879019)
--(axis cs:360,0.00467058484879019)
--(axis cs:370,0.00467058484879019)
--(axis cs:380,0.00467058484879019)
--(axis cs:390,0.00467058484879019)
--(axis cs:400,0.00467058484879019)
--(axis cs:400,0.000129415151209813)
--(axis cs:400,0.000129415151209813)
--(axis cs:390,0.000129415151209813)
--(axis cs:380,0.000129415151209813)
--(axis cs:370,0.000129415151209813)
--(axis cs:360,0.000129415151209813)
--(axis cs:350,0.000129415151209813)
--(axis cs:340,0.000129415151209813)
--(axis cs:330,0.00745669643947003)
--(axis cs:320,0.194655346559582)
--(axis cs:310,0.359188553842776)
--(axis cs:300,0.392559571899868)
--(axis cs:250,0.561143388516855)
--(axis cs:200,0.647569403923079)
--(axis cs:150,0.771760641267647)
--(axis cs:100,0.771635040088953)
--(axis cs:95,0.77730516616024)
--(axis cs:90,0.780686793998978)
--(axis cs:85,0.770068609928973)
--(axis cs:80,0.769508087363877)
--(axis cs:75,0.757132209759105)
--(axis cs:70,0.77398889210205)
--(axis cs:65,0.763980007993606)
--(axis cs:60,0.774446546654043)
--(axis cs:55,0.768266497463019)
--(axis cs:50,0.757428330766173)
--(axis cs:45,0.751919612728647)
--(axis cs:40,0.743547153835133)
--(axis cs:35,0.715657162781436)
--(axis cs:30,0.649683672937094)
--(axis cs:25,0.585957049022255)
--(axis cs:20,0.39895781046753)
--(axis cs:15,0.353121967875856)
--(axis cs:10,0.273627179818395)
--(axis cs:5,0.0946520918864213)
--(axis cs:4,0.00248775157689054)
--(axis cs:3,-0.000333053216343211)
--(axis cs:2,0.000969973673594152)
--(axis cs:1,-0.000388868236194652)
--cycle;

\addplot [ultra thick, darkorange2551490]
table {%
1 0.0026
2 0.0092
3 0.0076
4 0.0166
5 0.1558
10 0.3566
15 0.4286
20 0.5078
25 0.6506
30 0.7188
35 0.7498
40 0.7638
45 0.7712
50 0.7742
55 0.7814
60 0.795
65 0.789
70 0.7932
75 0.7868
80 0.7976
85 0.7998
90 0.8032
95 0.8058
100 0.798
150 0.7938
200 0.7046
250 0.6144
300 0.4426
310 0.3988
320 0.2936
330 0.0166
340 0.0024
350 0.0024
360 0.0024
370 0.0024
380 0.0024
390 0.0024
400 0.0024
};
\addplot [ultra thick, skyblue102204238, dashed]
table {%
333 -0.0421230533399924
333 0.876029018943558
};

\nextgroupplot[
width=8cm, 
height=5cm, 
legend cell align={left},
legend style={
  fill opacity=1.0,
  draw opacity=1,
  text opacity=1,
  at={(0.03,0.97)},
  anchor=north west,
  draw=lightgray204
},
tick pos=both,
xlabel={Minimum Number Effective Samples},
xmajorgrids,
xmin=-18.95, xmax=419.95,
xtick style={color=black},
ylabel={Distance Error [m]},
ymajorgrids,
ymin=0.124910715900248, ymax=0.892875450102703,
ytick style={color=black}
]
\path [fill=darkorange2551490, fill opacity=0.3]
(axis cs:1,0.159818203818541)
--(axis cs:1,0.782191448014668)
--(axis cs:2,0.606579090308787)
--(axis cs:3,0.733922636429517)
--(axis cs:4,0.588163269934282)
--(axis cs:5,0.373107220730002)
--(axis cs:10,0.314876252496882)
--(axis cs:15,0.297318191902544)
--(axis cs:20,0.264757479975651)
--(axis cs:25,0.224328549102495)
--(axis cs:30,0.220654213822848)
--(axis cs:35,0.212113589127987)
--(axis cs:40,0.200005799833891)
--(axis cs:45,0.204403955623556)
--(axis cs:50,0.206788523800562)
--(axis cs:55,0.209240130700299)
--(axis cs:60,0.201225959660073)
--(axis cs:65,0.206224626404019)
--(axis cs:70,0.204083891686413)
--(axis cs:75,0.210763884409825)
--(axis cs:80,0.205630260066336)
--(axis cs:85,0.203142784089019)
--(axis cs:90,0.216407252183632)
--(axis cs:95,0.205397705771031)
--(axis cs:100,0.207763739311105)
--(axis cs:150,0.233803009260222)
--(axis cs:200,0.256665064857435)
--(axis cs:250,0.297029214805569)
--(axis cs:300,0.344169542587653)
--(axis cs:310,0.337060382694344)
--(axis cs:320,0.426971090854234)
--(axis cs:330,0.648755275856382)
--(axis cs:340,0.85796796218441)
--(axis cs:350,0.85796796218441)
--(axis cs:360,0.857827290225282)
--(axis cs:370,0.85796796218441)
--(axis cs:380,0.85796796218441)
--(axis cs:390,0.85796796218441)
--(axis cs:400,0.85796796218441)
--(axis cs:400,0.477222455856579)
--(axis cs:400,0.477222455856579)
--(axis cs:390,0.477222455856579)
--(axis cs:380,0.477222455856579)
--(axis cs:370,0.477222455856579)
--(axis cs:360,0.472708958489269)
--(axis cs:350,0.477222455856579)
--(axis cs:340,0.477222455856579)
--(axis cs:330,0.480234591874419)
--(axis cs:320,0.307389654505195)
--(axis cs:310,0.303092607476776)
--(axis cs:300,0.298028822556086)
--(axis cs:250,0.265934931533008)
--(axis cs:200,0.240602871698074)
--(axis cs:150,0.211498010733892)
--(axis cs:100,0.195738952270647)
--(axis cs:95,0.192041498627079)
--(axis cs:90,0.187483563648713)
--(axis cs:85,0.190654112422686)
--(axis cs:80,0.184604073072359)
--(axis cs:75,0.185570237186093)
--(axis cs:70,0.183311263096318)
--(axis cs:65,0.182035640733946)
--(axis cs:60,0.186975124603224)
--(axis cs:55,0.185843087009598)
--(axis cs:50,0.185363959691121)
--(axis cs:45,0.185238408506678)
--(axis cs:40,0.181908188995115)
--(axis cs:35,0.182924268017552)
--(axis cs:30,0.185616893860836)
--(axis cs:25,0.192121190718809)
--(axis cs:20,0.222495025681366)
--(axis cs:15,0.24534526871061)
--(axis cs:10,0.263906314432524)
--(axis cs:5,0.293549760152424)
--(axis cs:4,0.240615105788975)
--(axis cs:3,0.254323845261734)
--(axis cs:2,0.278017550337629)
--(axis cs:1,0.159818203818541)
--cycle;

\addplot [ultra thick, darkorange2551490]
table {%
1 0.471004825916605
2 0.442298320323208
3 0.494123240845626
4 0.414389187861629
5 0.333328490441213
10 0.289391283464703
15 0.271331730306577
20 0.243626252828509
25 0.208224869910652
30 0.203135553841842
35 0.197518928572769
40 0.190956994414503
45 0.194821182065117
50 0.196076241745841
55 0.197541608854949
60 0.194100542131648
65 0.194130133568982
70 0.193697577391366
75 0.198167060797959
80 0.195117166569348
85 0.196898448255853
90 0.201945407916172
95 0.198719602199055
100 0.201751345790876
150 0.222650509997057
200 0.248633968277755
250 0.281482073169288
300 0.321099182571869
310 0.32007649508556
320 0.367180372679715
330 0.564494933865401
340 0.667595209020494
350 0.667595209020495
360 0.665268124357275
370 0.667595209020494
380 0.667595209020494
390 0.667595209020494
400 0.667595209020495
};
\addlegendentry{\small{ML-Cur}}
\addplot [ultra thick, skyblue102204238, dashed, forget plot]
table {%
333 0.124910715900248
333 0.892875450102703
};
\end{groupplot}

\end{tikzpicture}}
    \caption{\textbf{Table Tennis.} Model performance with 15 components, for an increasing number of minimum effective samples over ten random seeds and training-test selections.
    For small numbers of effective samples, the model specializes each component to very few training samples.
    The performance suffers as the context space is not appropriately covered.
    The dashed line indicates the point where the minimum number of components times the number of effective samples per component becomes larger than the total sample size.
    Approaching this threshold leads to the discussed degenerating behavior.}
    \label{fig:tt_alpha}
\end{figure}
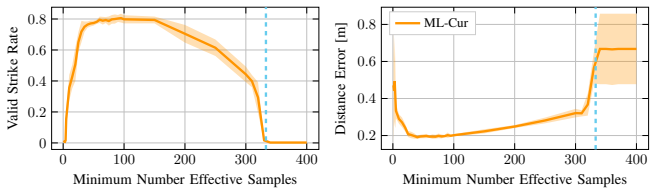

Additionally, for ML-Cur, we investigate the effect of changing the minimum number of effective samples to be covered by each component. 
We fix the number of MoE components at 15 and present the results in \autoref{fig:tt_alpha}. For small numbers of effective samples, the model is free to specialize each component to only very few training samples.
Thus, the performance suffers as the context space is not appropriately covered. For a large number of effective samples the curriculum weights become more uniform, leading to \textit{outlier-sensitivity}, \textit{locality-violations} and even \textit{mode-averaging}. Thus the MoE components are unable to specialize on samples and the overall model performance suffers.

\begin{figure}[t]
    \centering
    \resizebox{0.48\textwidth}{!}{\input{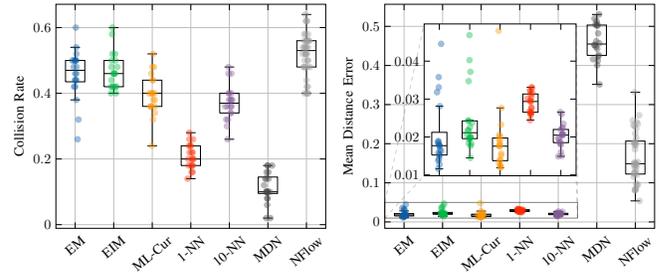}}
    \caption{\textbf{Obstacle Avoidance.} Distribution of test errors over 20 trials.
    MDN and NFlow can not solve the task. KNN ($k=1$) has the lowest collision rate, but the highest distance error. ML-Cur has the lowest distance error and second lowest collision rate, outperforming the other models. 
    }
    \label{fig:vo_boxplot}
\end{figure}

\textbf{Obstacle Avoidance.}
We conducted a real-world robot experiment with a Franka Emika Panda 7DoF Manipulator.
We collect versatile human expert demonstrations using teleoperation with a virtual twin: Moving a physical robot by hand and mirroring its state in a MuJoCo \cite{todorov2012mujoco}-based environment.
We place three static, cylindrical objects in the simulation environment and move the robot end-effector around these obstacles to the desired target position.
The task is contextualized by a 4D vector, describing the three obstacles and the goal's Y-axis position.
The X-coordinate is fixed for all four objects.
We train MoE models with twelve components using ML-Cur, EM, and EIM. As additional baselines, we include MDN, NFlow, and KNN with one (1-NN) and ten (10-NN) nearest neighbors. The models are trained on 228 collision-free demonstrations and tested on 50 random test contexts using MuJoCo.
We repeated this experiment 20 times with different random seeds and train-test selections and show the models' error distribution on the test data in Figure \ref{fig:vo_boxplot}.
As metrics, we consider the rate of generated trajectories, which collide with at least one obstacle, and the mean distance error (MDE) of the collision-free trajectories to the target position. MDN is unable to generate valid trajectories, resulting in the lowest collision rate, but the highest distance to the goal position. KNN with $k=1$ has a lower collision rate than ML-Cur, but slightly higher distance error. ML-Cur outperforms the remaining baselines.

\textbf{Ablation Studies.}
\begin{figure*}[t]
        \centering
        \begin{minipage}[t]{0.2\textwidth}
            \centering
            \subcaption{\small{ML-Cur}}\label{fig:ml_cur}
            \includegraphics[width=\textwidth]{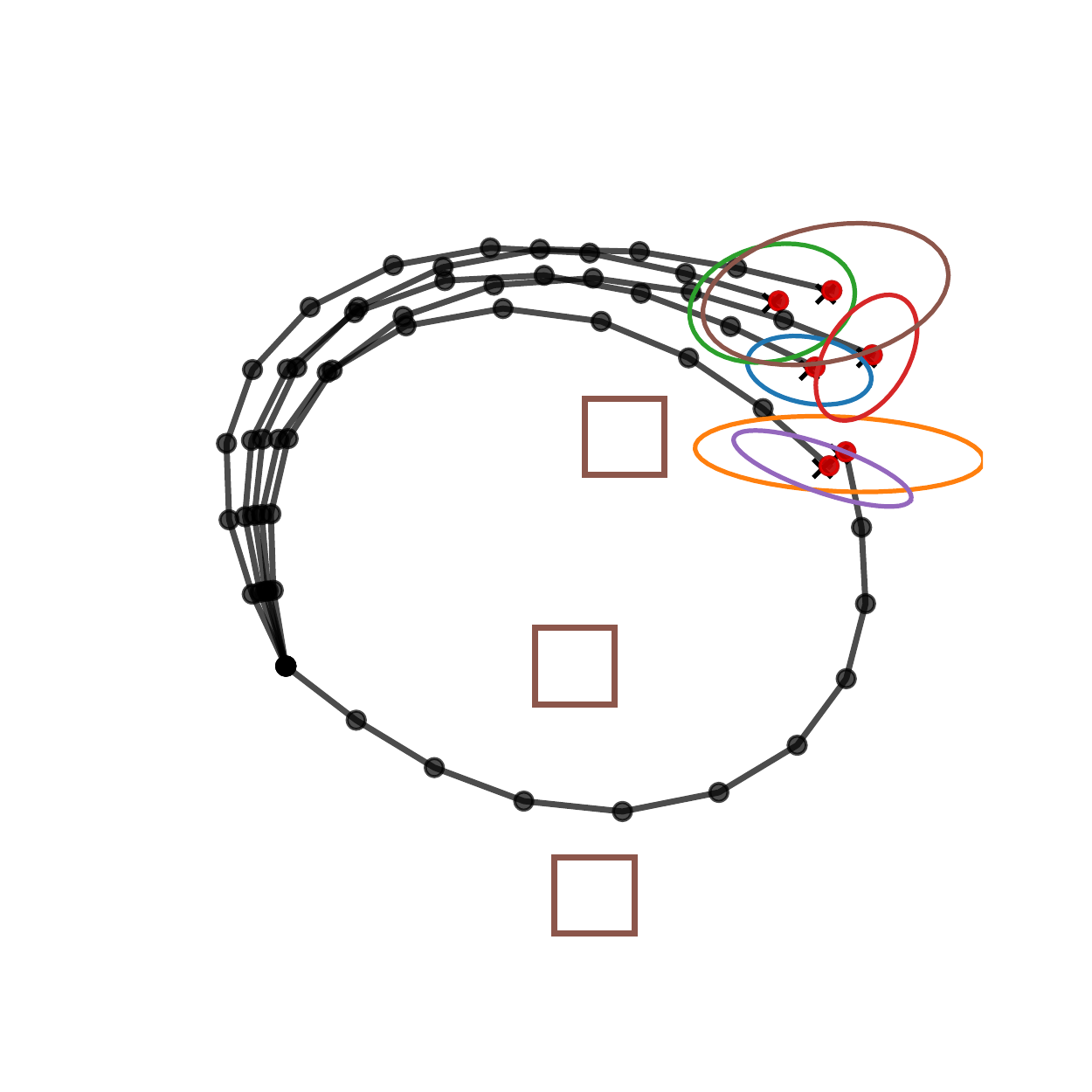}
        \end{minipage}%
        \hspace{0.015\textwidth}%
        \begin{minipage}[t]{0.2\textwidth}
            \centering 
            \subcaption{\small{No Data Weights}}\label{fig:weight}
            \includegraphics[width=\textwidth]{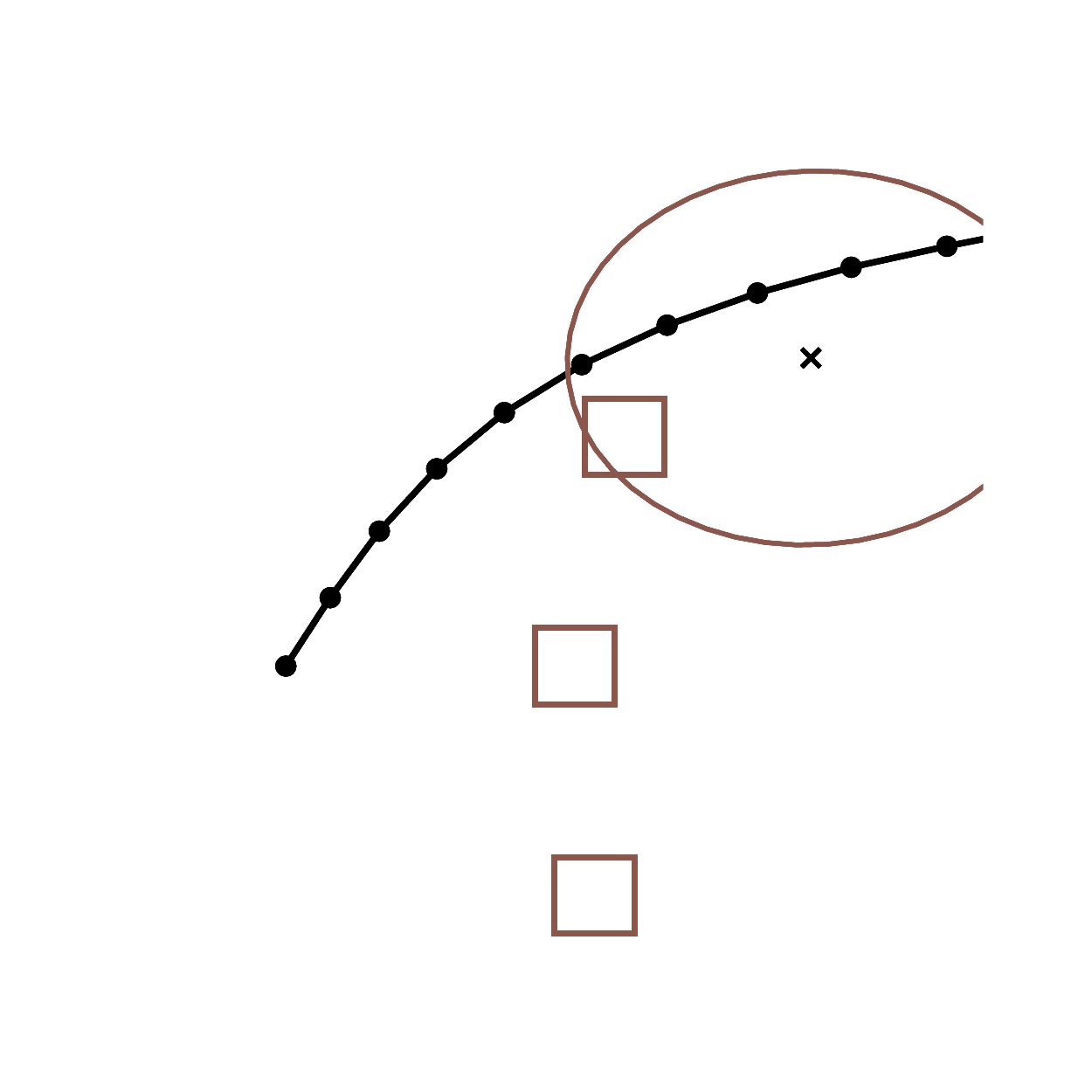}
        \end{minipage}%
        \hspace{0.015\textwidth}%
        \begin{minipage}[t]{0.2\textwidth}
            \centering 
            \subcaption{\small{With Locality Violation}}\label{fig:loc_vio}
            \includegraphics[width=\textwidth]{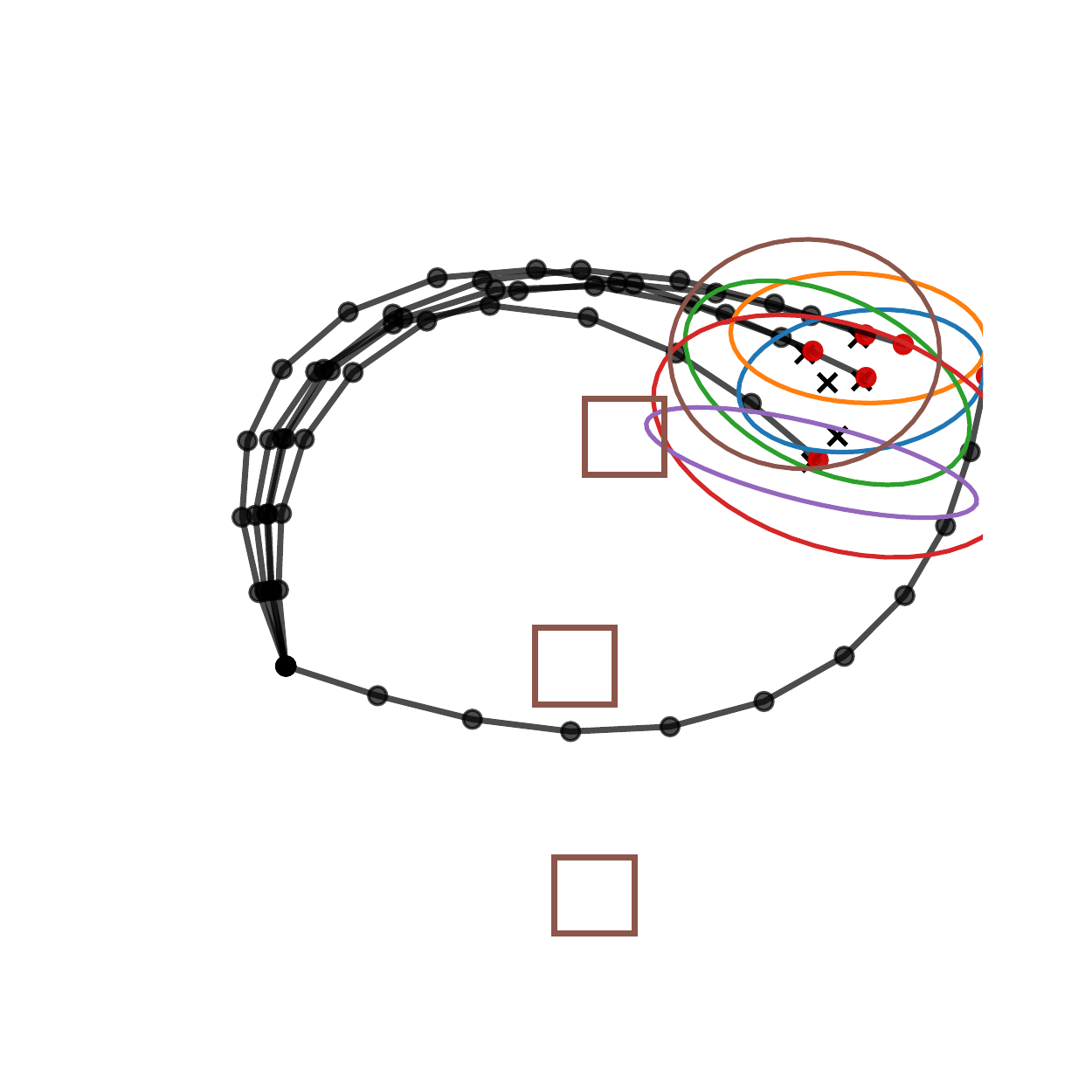}
        \end{minipage}%
        \hspace{0.015\textwidth}%
        \begin{minipage}[t]{0.2\textwidth}
        \centering 
        \subcaption{\small{Without Responsibility}}\label{fig:resp}
        \includegraphics[width=\textwidth]{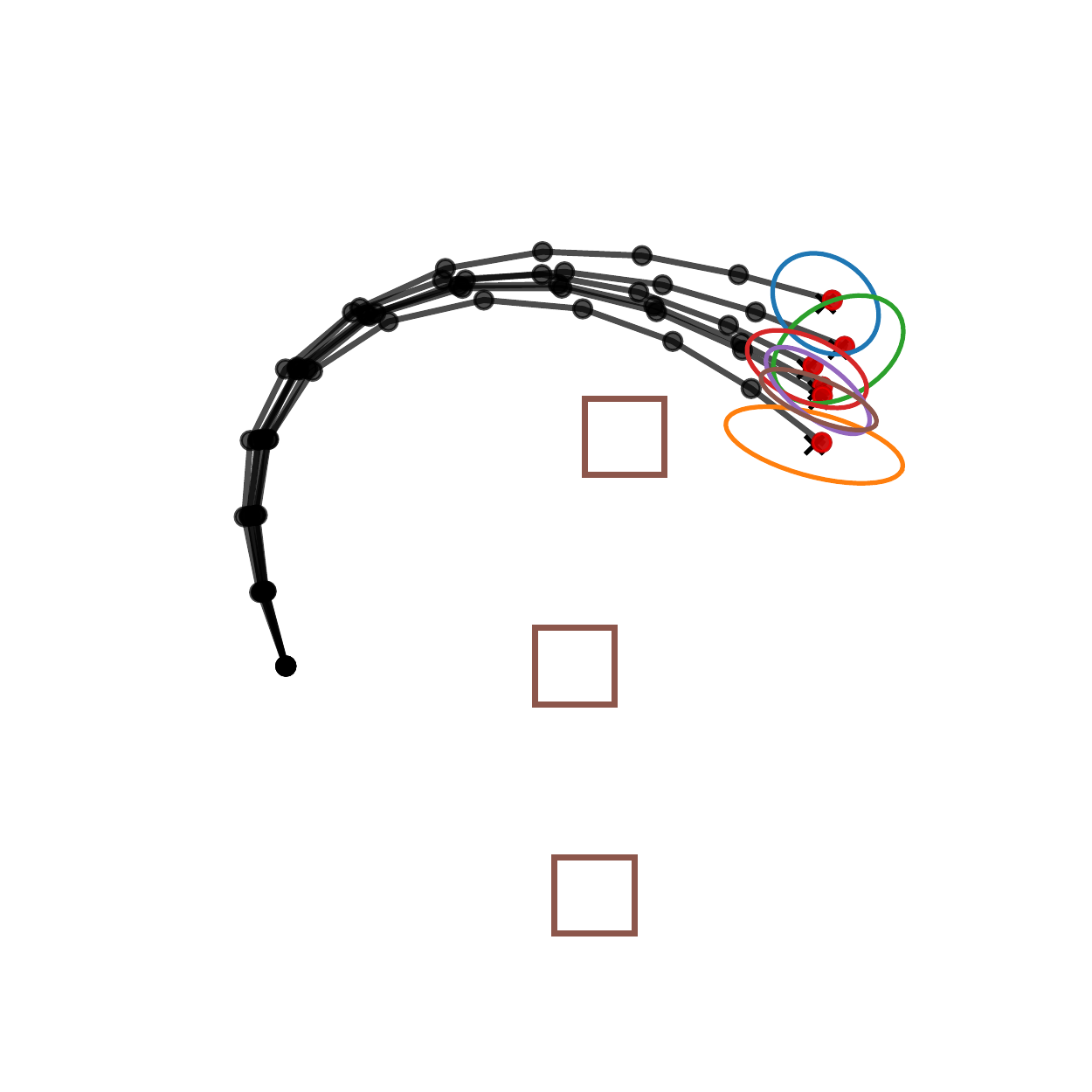}
        \end{minipage}
        \caption[]{
        The planar reacher configurations at the means of the context distributions. 
        The six Gaussian context distributions are shown as colored uncertainty ellipses. 
        \ref{fig:ml_cur} shows the result for ML-Cur. 
        \ref{fig:weight} demonstrates that removing the data weights reduces ML-Cur to a single expert model, averaging over the whole context space.
        \ref{fig:loc_vio} indicates that violating locality results in context components covering larger areas.
        \ref{fig:resp} shows that removing the responsibilities from the optimization results in heavily overlapping components, yielding less versatile solutions. 
        }
        \label{fig:ablation_vis}
    \end{figure*}
We perform several ablation studies in order to show the relative importance of the different terms of the ML-Cur objective. To that end, we use the table tennis task with $20$ components for a quantitative analysis and the planar reacher task for a qualitative analysis.

\textit{Data Weight Ablation.} We compare ML-Cur as proposed with a version which is optimized ignoring the data weights $\idatacompweight$ as introduced in \autoref{eq::LowerBound}. Consequently, we treat every data point as equally important. 

Ignoring the data weights lead to equal updates for all components and thus reduces the mixture of experts model to a single component. As a result, the performance degrades drastically on all experiments as shown in \autoref{tab:weights}. 
In \autoref{fig:weight} we visualize how the model behaves on the planar reaching task. Due to the lack of the weights, it can not concentrate on specific data points and hence can not adjust its curriculum which lead to a poor performance.

\begin{table}
    \caption{We ablate key aspects of our objective on the table tennis task. As the ablation study shows all the aspects contribute to the superior performance of ML-Cur.}
    \centering
    \begin{tabular}{ccc}
        \toprule 
        & \textbf{Valid Strikes} ($\uparrow$) & \textbf{Distance Error} ($\downarrow$) \\
        \midrule 
        ML-Cur & $\mathbf{0.791 \pm 0.034}$ & $\mathbf{0.201 \pm 0.007}$ \\ 
        \midrule
        No Data Weights &  $0.002 \pm 0.001$ & $0.62 \pm 0.218$ \\
        With Locality Violation &  $0.653 \pm 0.073$ & $0.209 \pm 0.013$ \\
        Without Responsibilities &  $0.667 \pm 0.077$ & $0.216 \pm 0.012$ \\
        \bottomrule
    \end{tabular}    
    \label{tab:weights}
\end{table}

\textit{Locality Violation Ablation.}  Next, we compare ML-Cur with a version which does not make use of the Gaussian context components $\ctxtcompdata$ during training. 
Without $\ctxtcompdata$ each expert might cover an arbitrary large or disjoint regions in the context space.
After training, we fit the Gaussian context components in order to be able to use the mixture model for inference. 
As a result, mode averaging leads to performance losses as shown in \autoref{tab:weights} and visualized in \autoref{fig:loc_vio}. 

\begin{figure}[t]
    \centering
    \resizebox{0.45\textwidth}{!}{\input{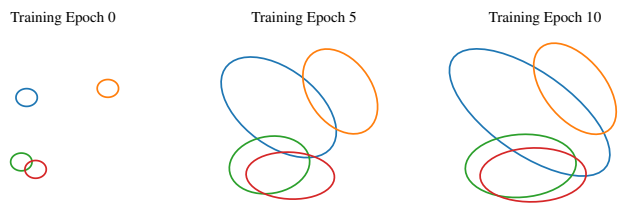}}
    \caption{
    The expansion of the Gaussian context distributions in the table tennis task show the adjustment of the curriculum.
    }
    \label{fig:curr_abl}
\end{figure}

\textit{Responsibility Ablation.} Lastly, we compare ML-Cur with a version where we do not maximize the data weight entropy $\mathcal{H}(\cvec \nu)$ and hence ignore the data weight responsibility $\idataresponsibility$ but instead maximize the per component entropy $\mathcal{H}(\cvec \nu_o)$. 
As a consequence, components are not punished when covering equal regions in the context space, leading to performance losses as shown in Table \ref{tab:weights} and visualized in \autoref{fig:resp}. \autoref{fig:curr_abl} shows the expansion of context components throughout training.

\section{Conclusion}
We introduced ML-Cur, a curriculum-based approach to versatile skill learning by imitation, using a re-parameterized MoE over movement primitives to model contextualized and versatile behavior.
ML-Cur solves three common problems of maximum likelihood learning for MoEs: \textit{outlier-sensitivity}, \textit{locality-violation} in context space and \textit{mode-averaging} in behavior space. We introduced a new objective that considers per-component datapoint weights and context distributions combined with an entropy bonus. The resulting objective yields expressive yet specialized components based on individually adjusted curricula. The variational lower bound allows for an efficient optimization while encouraging diversity between the components. 
The resulting method, ML-Cur, outperforms maximum likelihood approaches and even EIM in all conducted experiments, highlighting the advantages of our method.
One of our experiments uses versatile real-world human data, which suffers from the initially described issues.

\textbf{Limitations.} A limitation of our current approach is the linearity of the experts. 
While using a mixture of movement primitives still allows us to represent complex behavior it also limits our choice of the context space. Hence, the experiments in this paper do not consider high dimensional context, e.g., images or point clouds. 
However, for higher dimensional contexts, the linear representation would not be sufficient, and we would need to work with nonlinear experts.

\section*{Acknowledgements}
The research presented in this paper was funded by the Deutsche Forschungsgemeinschaft (German Research Foundation) – 448648559 and supported by the Carl Zeiss Foundation under the project JuBot (Jung Bleiben mit Robotern).
\bibliography{bibliography}  
\bibliographystyle{IEEEtranN}

\end{document}